\documentclass[10pt,twocolumn,letterpaper]{article}
\usepackage[utf8]{inputenc}
\usepackage{authblk}
\usepackage{cvpr}              % To produce the CAMERA-READY version
\usepackage{graphicx}
\usepackage{amsmath}
\usepackage{amssymb}
\usepackage{booktabs}
\usepackage{placeins}
\usepackage{wrapfig}
\usepackage[export]{adjustbox}
\usepackage{enumitem}
\usepackage{multirow}
\usepackage[accsupp]{axessibility} 
\usepackage{bbding}
\usepackage{pifont}
\usepackage{wasysym}
\usepackage[pagebackref,breaklinks,colorlinks]{hyperref}

\usepackage[capitalize]{cleveref}
\crefname{section}{Sec.}{Secs.}
\Crefname{section}{Section}{Sections}
\Crefname{table}{Table}{Tables}
\crefname{table}{Tab.}{Tabs.}

\def\cvprPaperID{1204} % *** Enter the CVPR Paper ID here
\def\confName{CVPR}
\def\confYear{2023}

\makeatletter
\renewcommand\AB@affilsepx{ \text{         }   \protect\Affilfont}
\makeatother

\begin{document}
\title{Zero-Shot Object Counting}
\author[1]{Jingyi Xu}
\author[2]{Hieu Le}
\author[1]{Vu Nguyen}
\author[3]{Viresh Ranjan\thanks{Work done prior to joining Amazon}}
\author[1]{Dimitris Samaras}
\affil[1]{Stony Brook University} 
\affil[2]{EPFL}
\affil[3]{Amazon}
 
\maketitle

%%%%%%%%% ABSTRACT
\begin{abstract}
Class-agnostic object counting aims to count object instances of an arbitrary class at test time. %It is challenging but also enables many potential applications. 
Current methods for this challenging problem require human-annotated exemplars as inputs, which are often unavailable for novel categories, especially for autonomous systems. Thus, we propose zero-shot object counting (ZSC), a new setting where only the class name is available during test time. Such a counting system does not require human annotators in the loop and can operate automatically. Starting from a class name, we propose a method that can accurately identify the optimal patches which can then be used as counting exemplars. Specifically, we first construct a class prototype to select the patches that are likely to contain the objects of interest, namely class-relevant patches. Furthermore, we introduce a model that can quantitatively measure how suitable an arbitrary patch is as a counting exemplar. By applying this model to all the candidate patches, we can select the most suitable patches as exemplars for counting. Experimental results on a recent class-agnostic counting dataset, FSC-147, validate the effectiveness of our method. Code is available at \url{https://github.com/cvlab-stonybrook/zero-shot-counting}.  %Our method can serve as a solid baseline and facilitate future work in zero-shot object counting.
\end{abstract}

%%%%%%%%% BODY TEXT
\section{Introduction}
\label{sec:intro}
\begin{figure}[t]
\begin{center}
% \begin{overpic} 
% [width=\linewidth]
% {example-image-a}
% \end{overpic}
\includegraphics[width=0.7\linewidth]{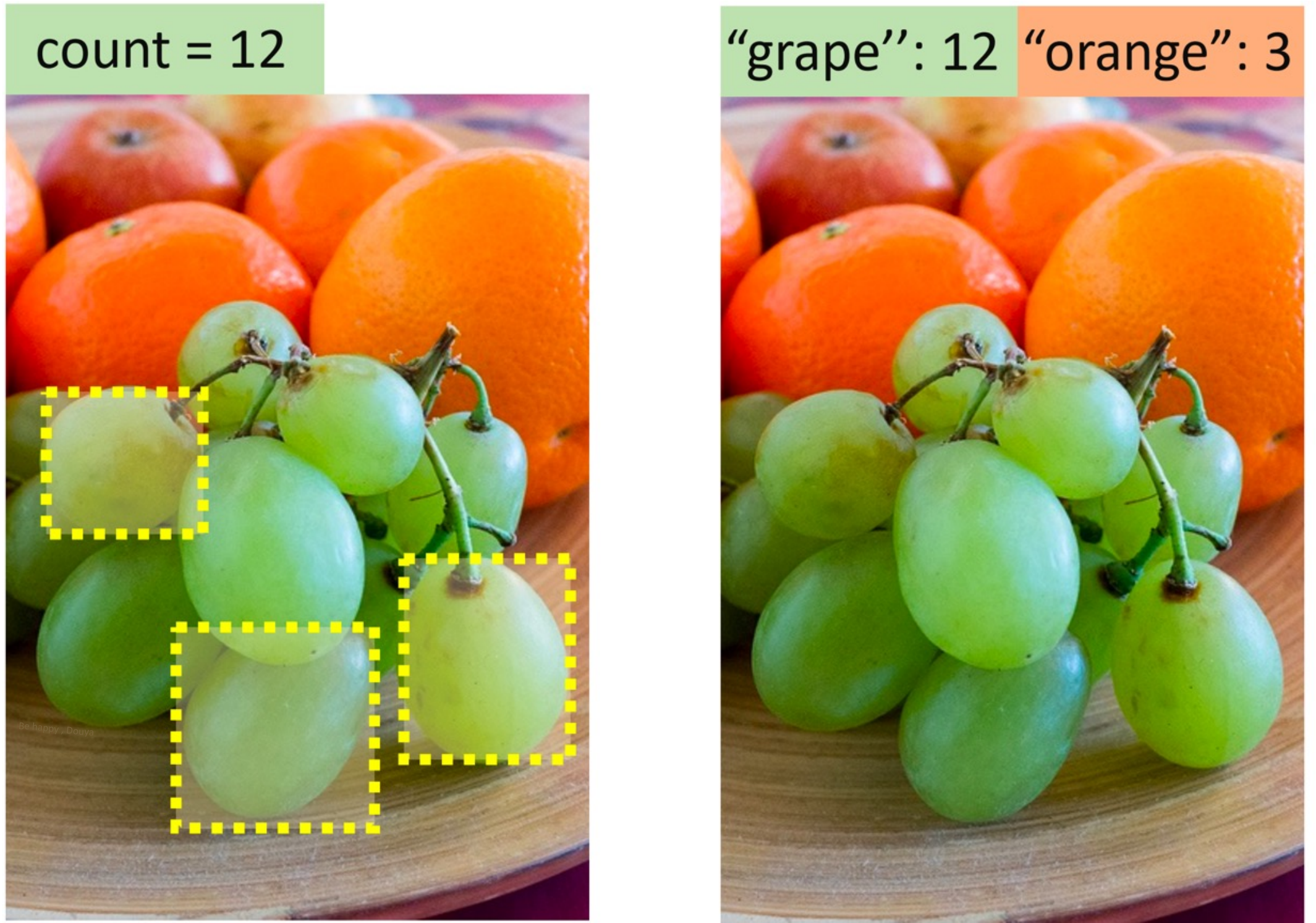}
\makebox[0.45\linewidth]{(a) Few-shot Counting }
\makebox[0.45\linewidth]{(b) Zero-Shot Counting}
\end{center} \vspace{-14pt}
\caption{
Our proposed task of zero-shot object counting (ZSC). Traditional few-shot counting methods require a few exemplars of the object category (a). We propose zero-shot counting where the counter only needs the class name to count the number of object instances. (b). Few-shot counting methods require human annotators at test time while zero-shot counters can be fully automatic.
}  \vspace{-14pt}
\label{fig:teaser} 
\end{figure}

Object counting aims to infer the number of objects in an image. Most of the existing methods focus on counting objects from specialized categories such as human crowds \cite{Sam2022SSCrowd}, cars \cite{Mundhenk2016ALC}, animals \cite{Arteta2016CountingIT}, and cells \cite{Xie2018MicroscopyCC}. These methods count only a single category at a time. 
Recently, class-agnostic counting \cite{Ranjan2021LearningTC,Shi2022SimiCounting,Lu2018CAC} has been proposed to count objects of arbitrary categories. Several human-annotated bounding boxes of objects are required to specify the objects of interest (see Figure \ref{fig:teaser}a). However, having humans in the loop is not practical for many real-world applications, such as fully automated wildlife monitoring systems or visual anomaly detection systems. 
%Therefore, exemplar-free class-agnostic counting \cite{Ranjan2022Exemplar} has been recently proposed as a useful task to eliminate the need of human-annotated inputs in the counting process.

\definecolor{azure(colorwheel)}{rgb}{0.0, 0.5, 1.0}
\definecolor{amber}{rgb}{1.0, 0.75, 0.0}

A more practical setting, exemplar-free class-agnostic counting, has been proposed recently by Ranjan \textit{et al.}\cite{Ranjan2022Exemplar}. They introduce RepRPN, which first identifies the objects that occur most frequently in the image, and then uses them as exemplars for object counting. Even though RepRPN does not require any annotated boxes at test time, the method simply counts objects from the class with the highest number of instances. Thus, it can not be used for counting a specific class of interest. The method is only suitable for counting images with a single dominant object class, which %significantly 
limits the potential applicability. %object counting applications.
 
Our goal is to build an exemplar-free object counter where we can specify what to count.
To this end, we introduce a new counting task in which the user only needs to provide the name of the class for counting rather than the exemplars (see Figure \ref{fig:teaser}b). In this way, the counting model can not only operate in an automatic manner but also allow the user to define what to count by simply providing the class name. Note that the class to count during test time can be arbitrary. For cases where the test class is completely unseen to the trained model, the counter needs to adapt to the unseen class without any annotated data. Hence, we name this setting zero-shot object counting (ZSC), inspired by previous zero-shot learning approaches \cite{Zheng2021ZeroShotIS,Bansal2018ZeroShotOD}. 
%As illustrated in Figure \ref{fig:teaser}, given an input image and a class label, a model is expected to infer the amount of instances of that specific class in the input image. Unlike the conventional class-agnostic counting task, our proposed ZSC task does not require any human-annotated exemplar as inputs, which is more challenging while practically useful in many scenarios.
% Talk about the main challenges of this task ... 

%Zero-shot counting requires the model to identify the exemplars for counting based on the given class label.   
%There are two main challenges for object counting using class name: (1) how to localize the area of the image that might contain the object of interests; (2) how to select the optimal patches within the area that can be used as exemplars for counting.

To count without any annotated exemplars, our idea is to identify a few patches in the input image containing the target object that can be used as counting exemplars. Here the challenges are twofold: 1) how to localize patches that contain the object of interest based on the provided class name, and 2) how to select \textit{good} exemplars for counting. Ideally, good object exemplars are visually representative for most instances in the image, which can benefit the object counter. In addition, we want to avoid selecting patches that contain irrelevant objects or backgrounds, which likely lead to incorrect object counts. 

To this end, we propose a two-step method that first localizes the class-relevant patches which contain the objects of interest based on the given class name, and then selects among these patches the optimal exemplars for counting. We use these selected exemplars,  together with a pre-trained exemplar-based counting model, to achieve exemplar-free object counting. %In essence, our method can automatically provide object exemplars based on the class name, which enables the conventional exemplar-based methods to a

In particular, to localize the patches containing the objects of interest, we first construct a class prototype in a pre-trained embedding space based on the given class name. %Inspired by previous zero-shot learning methods \cite{Xian2019ZeroShotLC}, 
To construct the class prototype, we train a conditional variational autoencoder (VAE) to generate features for an arbitrary class conditioned on its semantic embedding. The class prototype is computed by taking the average of the generated features. We then select the patches whose embeddings are the $k$-nearest neighbors of the class prototype as the class-relevant patches.

\begin{figure}[t]
\begin{center}
% \begin{overpic} 
% [width=\linewidth]
% {example-image-a}
% \end{overpic}
\includegraphics[width=\linewidth]{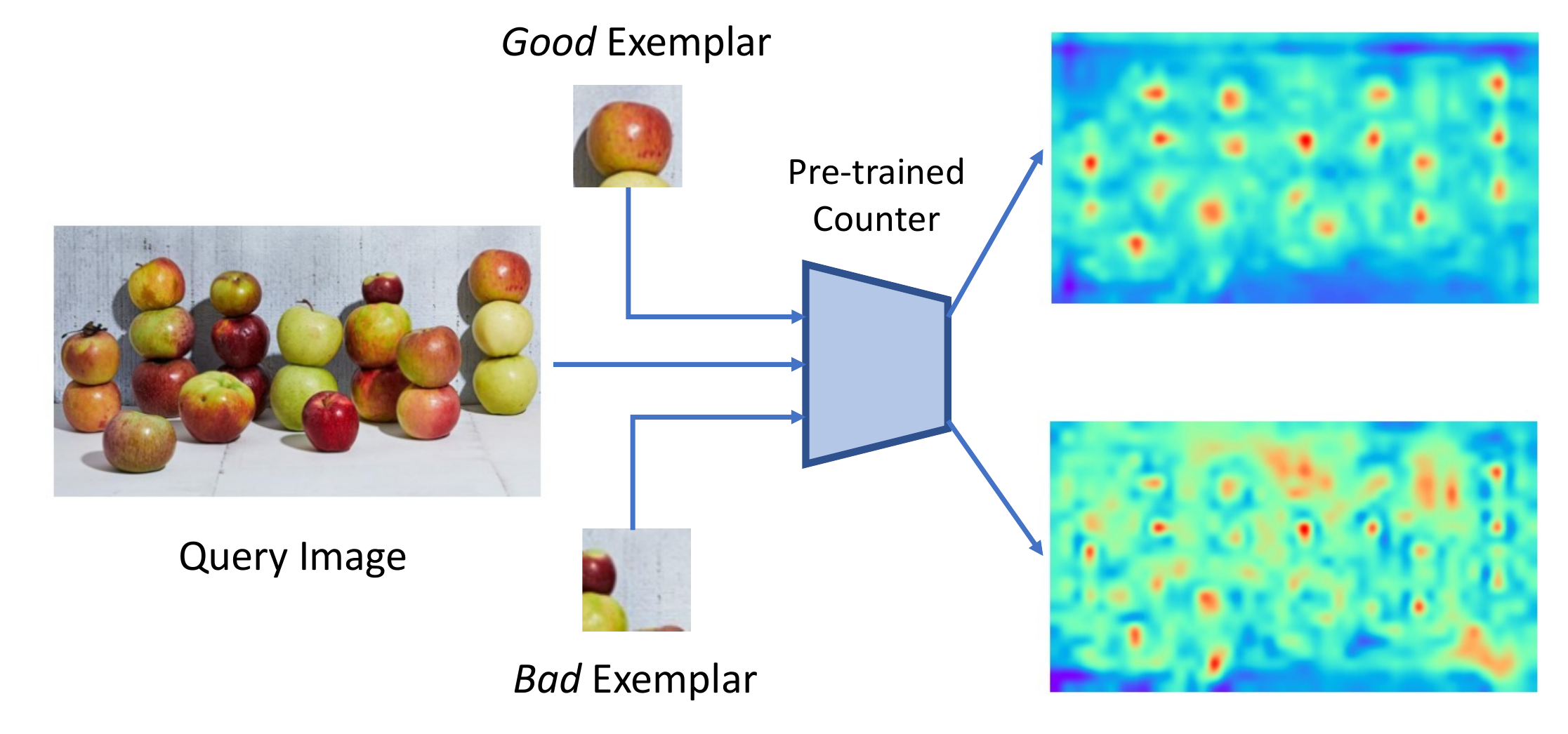} 
\end{center} \vspace{-4mm}
\caption{
Feature maps obtained using different exemplars given a pre-trained exemplar-based counting model. The feature maps obtained using good exemplars typically exhibit some repetitive patterns while the patterns from bad exemplars are more irregular.
} \vspace{-4mm}
\label{fig:teaser2}
\end{figure}

After obtaining the class-relevant patches, we further select among them the optimal patches to be used as counting exemplars. Here we observe that the feature maps obtained using \textit{good} 
exemplars and \textit{bad} exemplars often exhibit distinguishable differences. 
An example of the feature maps obtained with different exemplars is shown in Figure \ref{fig:teaser2}. The feature map from a \textit{good} exemplar typically exhibits some repetitive patterns (e.g., the dots on the feature map) that center around the object areas while the patterns from a \textit{bad} exemplar are more irregular and occur randomly across the image. Based on this observation, we train a model to measure the goodness of an input patch based on its corresponding feature maps. Specifically, given an arbitrary patch and a pre-trained exemplar-based object counter, we train this model to predict the counting error of the counter when using the patch as the exemplar. Here the counting error can indicate the goodness of the exemplar.
After this error predictor is trained, we  use it to select those patches with the smallest predicted errors as the final exemplars for counting.

Experiments on the FSC-147 dataset show that our method outperforms the previous exemplar-free counting method\cite{Ranjan2022Exemplar} by a large margin. We also provide analyses to show that patches selected by our method can be used in other exemplar-based counting methods to achieve exemplar-free counting. In short, our main contributions can be summarized as follows: 
\begin{itemize}%[leftmargin=*]
\setlength\itemsep{-.3em}
\item We introduce the task of zero-shot object counting that counts the number of instances of a specific class in the input image,  given only the class name and without relying on any human-annotated exemplars.
\item We propose a simple yet effective patch selection method that can accurately localize the optimal patches across the query image as exemplars for zero-shot object counting.
\item We verify the effectiveness of our method on the FSC-147 dataset, through extensive ablation studies and visualization results.% are provided to show the effectiveness of our method. 

\end{itemize}

\section{Related Work}

\begin{figure*}[!ht]
\begin{center}
\includegraphics[width=1.92\columnwidth]{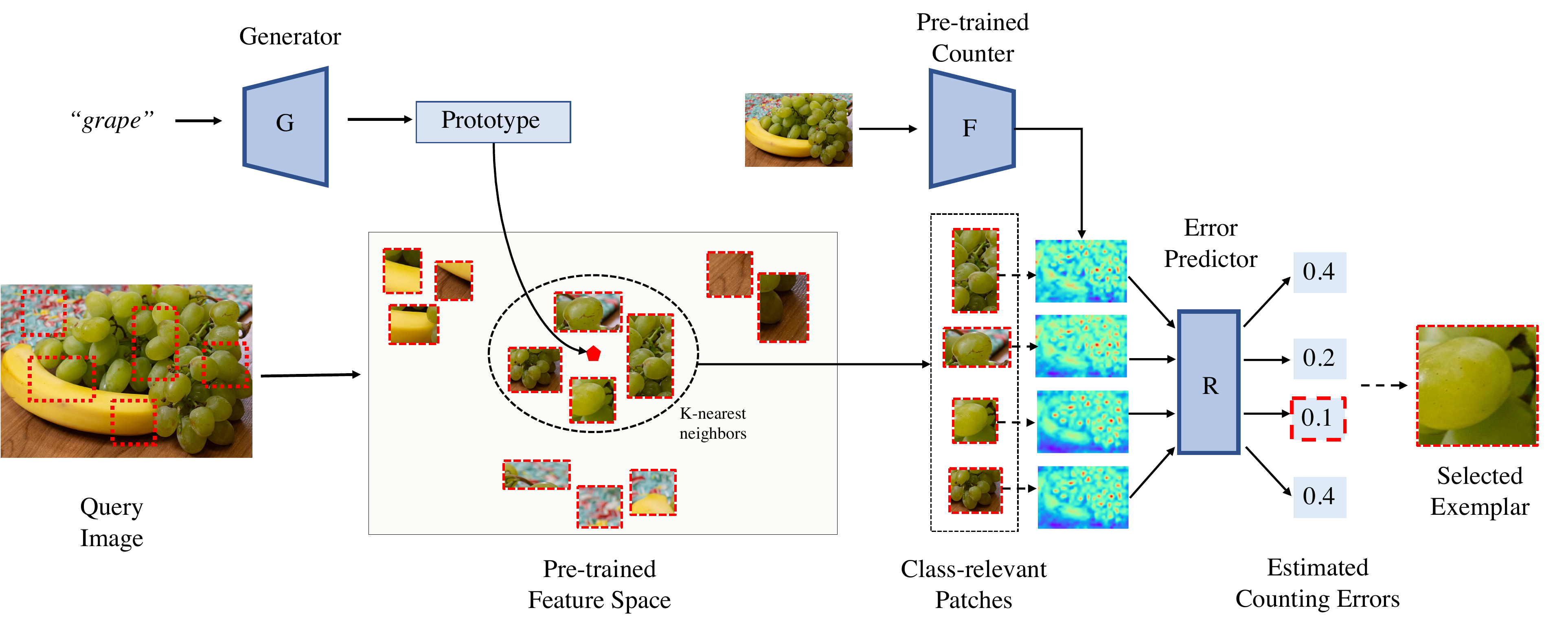} \vspace{-3mm}
\caption{
Overview of the proposed method. We first use a generative model to obtain a class prototype for the given class ({e.g.} grape) in a pre-trained feature space. Then given an input query image, we randomly sample a number of patches of various sizes and extract the corresponding feature embedding for each patch. We select the patches whose embeddings are the nearest neighbors of the class prototype as class-relevant patches. Then for each of the selected class-relevant patches, we use a pre-trained exemplar-based counting model to obtain the intermediate feature maps. Our proposed error predictor then takes the feature maps as input and predicts the counting error (here we use normalized counting errors). We select the patches with the smallest predicted errors as the final exemplar patches and use them for counting.
} \vspace{-6mm}
\label{fig:overview}
\end{center}

\end{figure*}

%\todo{read \cite{Yu2020EpisodeBasedPG}}

\label{sec:rw}
\subsection{Class-specific Object Counting} Class-specific object counting focuses on counting pre-defined categories, such as humans \cite{Lian2019DensityMR,zhang2016singleCC,Zhang2019AttentionalNF,Wang2021NWPUCrowdAL,Sindagi2019PushingTF,Idrees2018CompositionLF,Abousamra2021LocalizationIT,Zhang2015CrosssceneCC,Sam2022SSCrowd,Zhang2022CaliFree,Xiong2022DiscreteConstrainedRF,Liu2022LeveragingSF,Wan2021AGL}, animals \cite{Arteta2016CountingIT}, cells \cite{Xie2018MicroscopyCC}, or cars \cite{Mundhenk2016ALC,Hsieh2017DroneBasedOC}. Generally, existing methods can be categorized into two groups: detection-based methods \cite{Chattopadhyay2017CountingEO,Hsieh2017DroneBasedOC,Laradji2018WhereAT} and regression-based methods \cite{Zhang2015CrosssceneCC,Cholakkal2019ObjectCA,Cholakkal2022TowardsPS, Wang2020DMCrowd,zhang2016singleCC,Chan2008PrivacyPC, Liu2019ContextAwareCC}. Detection-based methods apply an object detector on the image and count the number of objects based on the detected boxes. Regression-based methods predict a density map for each input image, and the final result is obtained by summing up the pixel values. Both types of methods require abundant training data to learn a good model. Class-specific counters can perform well on trained categories. However, they can not be used to count objects of arbitrary categories at test time.

\subsection{Class-agnostic Object Counting} 
Class-agnostic object counting aims to count arbitrary categories given only a few exemplars \cite{Lu2018CAC,Ranjan2021LearningTC, Yang2021ClassagnosticFO,Shi2022SimiCounting,Gong2022ClassIntra,Nguyen2022fsoc,Liu2022CounTRTG,fsoc2023you,Arteta2014InteractiveOC}.
GMN \cite{Lu2018CAC} uses a shared embedding module to extract feature maps for both query images and exemplars, which are then concatenated and fed into a matching module to regress the object count.  %CFOCNet \cite{Yang2021ClassagnosticFO} convolves the feature maps from the exemplars over the query feature maps to obtain the density maps for object counting. 
FamNet \cite{Ranjan2021LearningTC} adopts a similar way to do correlation matching and further applies test-time adaptation. %Gong \textit{et al.} \cite{Gong2022ClassIntra} propose to use exemplar feature augmentation and edge matching to improve the counter's robustness against intra-class diversity.
These methods require human-annotated exemplars as inputs. Recently, Ranjan \textit{et al.} have proposed RepRPN \cite{Ranjan2022Exemplar}, which achieves exemplar-free counting by identifying exemplars from the most frequent objects via a Region Proposal Network (RPN)-based model. However, the class of interest can not be explicitly specified for the RepRPN. In comparison, our proposed method can count instances of a specific class given only the class name.

\subsection{Zero-shot Image Classification} 

Zero-shot classification aims to classify unseen categories for which data is not available during training \cite{Chen2018ZeroShotVR,Jayaraman2014ZeroshotRW,Frome2013DeViSEAD,Rezaei2020ZeroshotLA,RomeraParedes2015AnES,Atzmon2019AdaptiveCS,LeICCV2017,Le_2019_CVPR_Workshops,Le_2020_ECCV}. Semantic descriptors are mostly leveraged as a bridge to enable the knowledge transfer between seen and unseen classes. Earlier zero-shot learning (ZSL) works  
relate the semantic descriptors with visual features in an embedding space and recognize unseen samples by searching their nearest class-level semantic descriptor in this embedding space \cite{Lampert2009LearningTD,RomeraParedes2015AnES,Xian2016LatentEF,Zhang2017LearningAD}.
Recently, generative models \cite{JingyiICCV21,Xu2022GeneratingRS,m_Le-etal-ECCV18,le2020physicsbased} have been widely employed to synthesize unseen class data to facilitate ZSL \cite{Xian2019FVAEGAND2AF,Xian2019ZeroShotLC,Narayan2020LatentEF}. Xian \textit{et al.} \cite{Xian2019ZeroShotLC} use a conditional Wasserstein Generative Adversarial Network (GAN) \cite{Arjovsky2017WassersteinG} to generate unseen features which can then be used to train a discriminative classifier for ZSL.  
In our method, we also train a generative model conditioned on class-specific semantic embedding. Instead of using this generative model to  hallucinate data, we use it to compute a prototype for each class. This class prototype is then used to select patches that contain objects of interest.
\section{Method}
\label{sec:method}
%We briefly introduce the definition of few-shot learning problems in Section ~\ref{sec:definition}. We provide a rundown for our method in Section ~\ref{sec:overall_pipeline}.  

Figure \ref{fig:overview} summarizes our proposed method. Given an input query image and a class label, we first use a generative model to construct a class prototype for the given class in a pre-trained feature space. We then randomly sample a number of patches of various sizes and extract the feature embedding for each patch. The class-relevant patches are  those patches whose embeddings are the nearest neighbors of the class prototype in the embedding space. %We further apply an error predictor on these patches and select the patches with the smallest predicted errors as the final exemplars. 
We further use an error predictor to select the patches with the smallest predicted errors as the final exemplars for counting.
%select the patches that are likely to contain the objects of interest, namely class-relevant patches. Then we select among these class-relevant patches the most suitable patches for counting. 
We use the selected exemplars in an exemplar-based object counter to infer the object counts. For the rest of the paper, we denote this exemplar-based counter  as the ``base counting model".
%Our method requires a pre-trained exemplar-based counting model. 
We will first describe how we train this base counting model and then present the details of our  patch selection method.
%
%We first train a base counting model with abundant training data with annotations. 
%This counting model can be used for counting given a query image and a few exemplars. Then we introduce our method to achieve zero-shot counting based on this pre-trained counting model.

\subsection{Training Base Counting Model} 
\label{sec:baseline_training}

We train our base counting model using abundant training images with annotations. Similar to previous works \cite{Ranjan2021LearningTC,Shi2022SimiCounting}, the base counting model uses the input image and the exemplars to obtain a density map for object counting. The model consists of a feature extractor $F$ and a counter $C$. Given a query image $I$ and an exemplar $B$ of an arbitrary class $c$, we input $I$ and $B$ to the feature extractor to obtain the corresponding output, denoted as $F(I)$ and $F(B)$ respectively. $F(I)$ is a feature map of size $d * h_I * w_I $ and $F(B)$ is a feature map of size $d * h_B * w_B $. We further perform global average pooling on $F(B)$ to form a feature vector $b$ of $d$ dimensions.

After feature extraction, we  obtain the similarity map $S$ by correlating the exemplar feature vector $b$ with the image feature map $F(I)$. 
Specifically, if $w_{ij} = F_{ij}(I)$ is the channel feature at spatial position $(i,j)$, $S$ can be computed by:
 \begin{equation}\label{eq:simi}
     S_{ij}(I, B) = w_{ij}^T b.
\end{equation}

In the case where $n$ exemplars are given, we  use Eq. \ref{eq:simi} to calculate $n$ similarity maps, and the final similarity map is the average of these $n$ similarity maps.

We then concatenate the image feature map $F(I)$ with the similarity map $S$, and input them into the counter $C$ to predict a density map $D$.
%The counter receives the channel-wise concatenation of the image feature map $F(X)$ and the similarity map $S$, and then predicts a density map $D_{pr}$. 
The final predicted count ${N}$ is obtained by summing over the predicted density map ${D}$:
\begin{equation} \label{eq:final_count}
 {N} = \sum_{i,j}D_{(i,j)}, \vspace{-2mm}   
\end{equation}
where ${D}_{(i,j)}$ denotes the density value for pixel $(i,j)$. 
%We adopt a conventional $l_2$ loss as the counting loss $L_\textnormal{count}$ to supervise the training of the counting model: 
The supervision signal for training the counting model is the $L_2$ loss between the predicted density map and the ground truth density map:

\begin{equation}\label{eq:counting_loss}
L_{\textnormal{count}} = \|D(I, B) - D^{*}(I)\|_2^2, 
\end{equation}
where $D^{*}$ denotes the ground truth density map.

\subsection{Zero-shot Object Counting}
%After being trained with abundant annotated data, the base model can be used for counting given a few human-annotated exemplars. 
In this section, we describe how we count objects of any unseen category given only the class name without access to any exemplar. Our strategy is to select a few patches in the image that can be used as exemplars for the base counting model. %Specifically, we first select the patches that are likely to contain objects of the target class, namely class-relevant patches. Then we select among these class-relevant patches the most suitable patches for counting. 
These patches are selected such that: 1) they contain the objects that we are counting and 2) they benefit the counting model, i.e., lead to small counting errors.
\subsubsection{Selecting Class-relevant Patches}
 %We aim to select the patches that are likely to contain the objects of the target class given the class name. To do this, 
 To select patches that contain the objects of interest, we first generate a class prototype based on the given class name using a conditional VAE model. 
 Then we randomly sample a number of patches across the query image and select the class-relevant patches based on the generated prototype.
%
%In the following, we first illustrate how we train the generative model to generate the class prototype. Then we describe how we select the class-relevant patches across the query image based on this generated prototype.

 \textbf{Class prototype generation.} 
 Inspired by previous zero-shot learning approaches \cite{Xian2019FVAEGAND2AF,Xian2019ZeroShotLC}, we train a conditional VAE model to generate features for an arbitrary class based on the semantic embedding of the class. The semantic embedding is obtained from a pre-trained text-vision model \cite{Radford2021LearningTV} given the corresponding class name.
 %The generative model is built based on a conditional variational autoencoder (VAE) framework \cite{vae}. 
 Specifically, we train the VAE model to reconstruct features in a pre-trained ImageNet feature space. The VAE is composed of an Encoder $E$, which maps a visual feature $x$ to a latent code $z$, and a decoder $G$ which reconstructs $x$ from $z$. Both $E$ and $G$ are conditioned on the semantic embedding $a$ .%, which is obtained from a pre-trained NLP model given the class name of the input $x$.
 The loss function for training this VAE for an input feature $x$ can be defined as:
 \begin{equation}\label{eq:cvae}
 \begin{aligned}
      L_{V}(x) =  \textnormal{KL} \left( q(z|x,a)||p(z|a) \right)  \\
      - \textnormal{E}_{q(z|x, a)}[\textnormal{log }p(x|z,a)].
\end{aligned}
\end{equation}

The first term is the Kullback-Leibler divergence between the VAE posterior $q(z|x,a)$ and a prior distribution $p(z|a)$. The second term is the decoder's reconstruction error. $q(z|x,a)$ is modeled as $E(x, a)$ and $p(x|z,a)$ is equal to $G(z, a)$. The prior distribution is assumed to be $\mathcal{N}(0,I)$ for all classes.

We can use the trained VAE to generate the class prototype for an arbitrary target class for counting. Specifically, given the target class name $y$, we first generate a set of features by inputting the respective semantic vector $a^y$ and a noise vector $z$ to the decoder $G$:
\begin{equation}
\mathbb{G}^y = \{ \hat{x} | \hat{x} = G(z, y), z \sim \mathcal{N}(0, I)\}.
\end{equation}
The class prototype $\textnormal{p}^y$ is computed by taking the mean of all the features generated by VAE:  
 \begin{equation}
 \label{eq:prototype}
 \textnormal{p}^y = \frac{1}{|\mathbb{G}^y|} {\sum}_{\hat{x} \in \mathbb{G}^y}
{\hat{x}} 
\end{equation}

\textbf{Class-relevant patch selection.} The generated class prototype can be considered as a class center representing the distribution of features of the corresponding class in the embedding space. Using the class prototype, we can select the class-relevant patches across the query image. Specifically, we first randomly sample $M$ patches of various sizes $\{b_1, b_2, ... , b_m\}$ across the query image and extract their corresponding ImageNet features $\{f_1, f_2, ... , f_m\}$. To select the class-relevant patches, we calculate the $L_2$ distance between the class prototype and the patch embedding, namely $d_i = \| f_i - \text{p}^y\|_2$. Then we select the patches whose embeddings are the $k$-nearest neighbors of the class prototype as the class-relevant patches.
Since the ImageNet feature space is highly discriminative, {i.e.}, features close to each other typically belong to the same class, the selected patches are likely to contain the objects of the target class.

\subsubsection{Selecting Exemplars for Counting}
Given a set of class-relevant patches and a pre-trained exemplar-based object counter, we aim to select a few exemplars from these patches that are optimal for counting. To do so, we introduce an error prediction network that predicts the counting error of an arbitrary patch when the patch is used as the exemplar. The counting error is calculated from the pre-trained counting model. %In essence, our proposed network measures how an arbitrary patch benefit the pre-trained counter.
%After obtaining the class-relevant patches, we further employ an error predictor to select among the class-relevant patches that most suitable patches for counting. 
% Here the optimal patches are ones that minimize the counting errors. 
%Here the optimal patches refer to the patches that will yield the smallest counting errors when being used as exemplars.
%The proposed error predictor is trained to predict the counting error of an arbitrary patch when the patch is used as the exemplar. The counting error is calculated from the pre-trained counting model. %After we train this error predictor, we use it to infer the counting errors for all candidate patches and select those with the smallest errors as the optimal patches for counting.
Specifically, to train this error predictor, given a query image $\bar{I}$ and an arbitrary patch $\bar{B}$ cropped from $\bar{I}$, we first use the base counting model to get the image feature map $F(\bar{I})$, similarity map $\bar{S}$, and the final predicted density map $\bar{D}$. The counting error of the base counting model can be written as:  
\begin{equation}
\label{eq:error}
%\epsilon = | \sum_{i,j} \bar{D}_{(i,j)} - \sum_{i,j} D^{*}_{(i,j)}|.
\epsilon = | \sum_{i,j} \bar{D}_{(i,j)} - \bar{N^*}|, \vspace{-2mm}
\end{equation}
where $\bar{N^*}$ denotes the ground truth object count in image $\bar{I}$. 
%where $N(\bar{I}, \bar{B})$ is the predicted final count computed using Eq. \ref{eq:final_count}, and $N^{*}(\bar{I})$ is the ground truth object count for image $\bar{I}$. 
$\epsilon$ can be used to measure the goodness of $\bar{B}$ as an exemplar for $\bar{I}$, {i.e.}, a small $\epsilon$ indicates that $\bar{B}$ is a suitable exemplar for counting and vice versa.

The error predictor $R$ is trained to regress the counting error produced by the base counting model. The input of $R$ is the channel-wise concatenation of the image feature map $F(\bar{I})$ and the similarity map $\bar{S}$. The training objective is the minimization of the mean squared error between the output of the predictor $R(F(\bar{I}), \bar{S})$ and the actual counting error produced by the base counting model $\epsilon$.

After the error predictor is trained, we can use it to select the optimal patches for counting. The candidates for selection here are the class-relevant patches selected by the class prototype in the previous step. For each candidate patch, we use the trained error predictor to infer the counting error when it is being used as the exemplar. The final selected patches for counting are the patches that yield the top-$s$ smallest counting errors.

\subsubsection{Using the Selected Patches as Exemplars}
Using the error predictor, we predict the error for each candidate patch and select the patches that lead to the smallest counting errors. The selected patches can then be used as exemplars for the base counting model to get the density map and the final count. We also conduct experiments to show that these selected patches can serve as exemplars for other exemplar-based counting models to achieve exemplar-free class-agnostic counting.     

\section{Experiments}
\label{sec:exp}

\subsection{Implementation Details}

\textbf{Network architecture} 
For the \textit{base counting model}, we use ResNet-50 as the backbone of the feature extractor, initialized with the weights of a pre-trained ImageNet model. The backbone outputs feature maps of $1024$ channels. For each query image, the number of channels is reduced to $256$ using an $1 \times 1$ convolution. For each exemplar, the feature maps are first processed with global average pooling and then linearly mapped to obtain a $256$-d feature vector. The counter consists of $5$ convolutional and bilinear upsampling layers to regress a density map of the same size as the query image. 
For the \textit{feature generation model}, both the encoder and the decoder are two-layer fully-connected (FC) networks with 4096 hidden units. LeakyReLU and ReLU are the non-linear activation functions in the hidden and output layers, respectively. The dimensions of the latent space and the semantic embeddings are both set to be $512$.
For the \textit{error predictor}, $5$ convolutional and bilinear upsampling layers are followed by a linear layer to output the counting error.

\textbf{Dataset} We use the FSC-147 dataset \cite{Ranjan2021LearningTC} to train the base counting model and the error predictor. FSC-147 is the first large-scale dataset for class-agnostic counting. It includes $6135$ images from $147$ categories varying from animals, kitchen utensils, to vehicles. The categories in the training, validation, and test sets do not overlap. The feature generator is trained on the MS-COCO detection dataset. Note that the previous exemplar-free method \cite{Ranjan2022Exemplar} also uses MS-COCO to pre-train their counter.
\begin{table*}[!h] 
  \centering
\resizebox{0.72\textwidth}{!}{%
  \begin{tabular}{l|c|cccc|cccc}
    \toprule
   \multirow{2}{*}{Method} & \multirow{2}{*}{Exemplars} & \multicolumn{4}{c|}{Val Set} & \multicolumn{4}{c}{Test Set} \\
    & & MAE & RMSE & NAE & SRE & MAE & RMSE & NAE & SRE \\
    \midrule
    \multirow{2}{*}{GMN \cite{Lu2018CAC}} & GT & {29.66} & {89.81} & - & - & {26.52} & {124.57} & - & - \\
    & RPN & {40.96} & {108.47} & - & - & {39.72} & {142.81} & - & - \\
    \midrule
    %\multirow{2}{*} {FamNet} & GT & {24.32} & {70.94} & {22.56} & {101.54} \\
    % & RPN & {48.62} & {120.77} & {53.82} & {155.38}  \\
    %\midrule
    \multirow{2}{*}{FamNet+ \cite{Ranjan2021LearningTC}} & GT & {23.75} & {69.07} & 0.52 & 4.25 & {22.08} & {99.54} & 0.44 & 6.45 \\
     & RPN & {42.85} & {121.59} & 0.75 & 6.94 & {42.70} & {146.08}  & 0.74 & 7.14 \\
     \midrule
     \multirow{2}{*}{BMNet \cite{Shi2022SimiCounting}} & GT & {19.06} & {67.95} & 0.26 & 4.39 & {16.71} & {103.31} & 0.26 & 3.32 \\
     & RPN & {37.26} & {108.54} & {0.42} & {5.43} & 37.22 & 143.13 & 0.41 & 5.31 \\
     \midrule
    \multirow{2}{*} {BMNet+ \cite{Shi2022SimiCounting}} & GT & {15.74} & {58.53} & {0.27} & {6.57} & 14.62 & 91.83 & 0.25 & 2.74 \\
     & RPN & {35.15} & {106.07} & {0.41} & {5.28} & 34.52 & 132.64 & 0.39 & 5.26 \\
    \midrule
    %{FamNet} & {32.15} & {98.75} & {32.27} & {131.46} \\
    %{BMNet} & & {34.09} & {105.05} & {31.27} & {124.85} \\
    %{BMNet+} & & {35.15} & {106.07} & {34.52} & {132.64} \\
    %{BMNet+} & {23.01} & {72.61} & {22.43} & {128.12} \\
    {RepRPN-Counter} \cite{Ranjan2022Exemplar} & - & {30.40} & {98.73} & - & - & {27.45} & {129.69} & - & -\\
    \midrule
    \multirow{3}{*}{Ours (Base)} & GT & {18.55} & {61.12} & 0.30 & 3.18 & {20.68} & {109.14} & 0.36 & 7.63 \\
     & RPN & 32.19 & 99.21 & 0.38 & 4.80 & 29.25 & 130.65 & 0.35 & 4.35  \\
     & Patch-Selection & \textbf{26.93} & \textbf{88.63} & \textbf{0.36} & \textbf{4.26} & \textbf{22.09} & \textbf{115.17} & \textbf{0.34} & \textbf{3.74} \\
    \bottomrule
  \end{tabular}} \\ \vspace{-5pt}
  \caption{ Quantitative comparisons on the FSC-147 dataset. ``GT" denotes using human-annotated boxes as exemplars. ``RPN" denotes using the top-3 RPN proposals with the highest objectness scores as exemplars. ``Patch-Selection" denotes using our selected patches as exemplars.
  }\label{tab:test_val}%
  \vspace{-9pt}
\end{table*}

\textbf{Training details} 
Both the base counting model and the error predictor are trained using the AdamW optimizer with a fixed learning rate of $10^{-5}$. The base counting model is trained for $300$ epochs with a batch size of $8$. We resize the input query image to a fixed height of $384$, and the width is adjusted accordingly to preserve the aspect ratio of the original image.
Exemplars are resized to $128 \times 128$ before being input into the feature extractor. 
The feature generation model is trained using the Adam optimizer and the learning rate is set to be $10^{-4}$. The semantic embeddings are extracted from CLIP \cite{Radford2021LearningTV}. To select the class-relevant patches, we randomly sample $450$ boxes of various sizes across the input query image and select $10$ patches whose embeddings are the $10$-nearest neighbors of the class prototype. The final selected patches are those that yield the top-$3$ smallest counting errors predicted by the error predictor.

\subsection{Evaluation Metrics}
We use Mean Average Error (MAE) and Root Mean Squared Error (RMSE) to measure the performance of different object counters. Besides, we follow \cite{Nguyen2022fsoc} to report the Normalized Relative Error (NAE) and Squared Relative Error (SRE). In particular, MAE = $\frac{1}{n} \sum_{i=1}^n |y_i-\hat{y_i}|$; RMSE = $\sqrt{\frac{1}{n} \sum_{i=1}^n (y_i-\hat{y_i})^2}$; NAE = $\frac{1}{n} \sum_{i=1}^n \frac{|y_i-\hat{y_i}|}{y_i}$; SRE = $\sqrt{\frac{1}{n} \sum_{i=1}^n \frac{(y_i-\hat{y_i})^2}{y_i}}$ where $n$ is the number of test images, and $y_i$ and $\hat{y_i}$ are the ground truth and the predicted number of objects for image $i$ respectively. %Compared with the absolute errors MAE and RMSE, the relative errors NAE and SRE better reflect the practical usage of visual counting \cite{Nguyen2022fsoc}.

\subsection{Comparing Methods}
We compare our method with the previous works on class-agnostic counting. RepRPN-Counter \cite{Ranjan2022Exemplar} is the only previous class-agnostic counting method that does not require human-annotated exemplars as input. In order to make other exemplar based class-agnostic methods including GMN (General Matching Network \cite{Lu2018CAC}), FamNet (Few-shot adaptation and matching Network \cite{Ranjan2021LearningTC}) and BMNet (Bilinear Matching Network \cite{Shi2022SimiCounting})  work in the exemplar-free setup, we replace the human-provided exemplars with the exemplars generated by a pre-trained object detector. Specifically, we use the RPN of Faster RCNN pre-trained on MS-COCO dataset and select the top-$3$ proposals with the highest objectness score as the exemplars. We also include the performance of these methods using human-annotated exemplars for a complete comparison.

\subsection{Results} 

\textbf{Quantitative results.} As shown in Table \ref{tab:test_val}, our proposed method outperforms the previous exemplar-free counting method \cite{Ranjan2022Exemplar} by a large margin, resulting in a reduction of $10.10$ \textit{w.r.t.} the validation RMSE and $14.52$ \textit{w.r.t.} the test RMSE. We also notice that the performance of all exemplar-based counting methods drops significantly when replacing human-annotated exemplars with RPN generated proposals. The state-of-the-art exemplar-based method BMNet+ \cite{Shi2022SimiCounting}, for example, shows an $19.90$ error increase \textit{w.r.t.} the test MAE and a $40.81$ increase \textit{w.r.t.} the test RMSE. In comparison, the performance gap is much smaller when using our selected patches as exemplars, as reflected by a $1.41$ increase \textit{w.r.t.} the test MAE and a $6.03$ increase \textit{w.r.t.} the test RMSE. Noticeably, the NAE and the SRE on the test set are even reduced when using our selected patches compared with the human-annotated exemplars.

\begin{figure*}[ht!]
 \centering
\hspace*{-0.1cm}\includegraphics[width=0.75\linewidth]{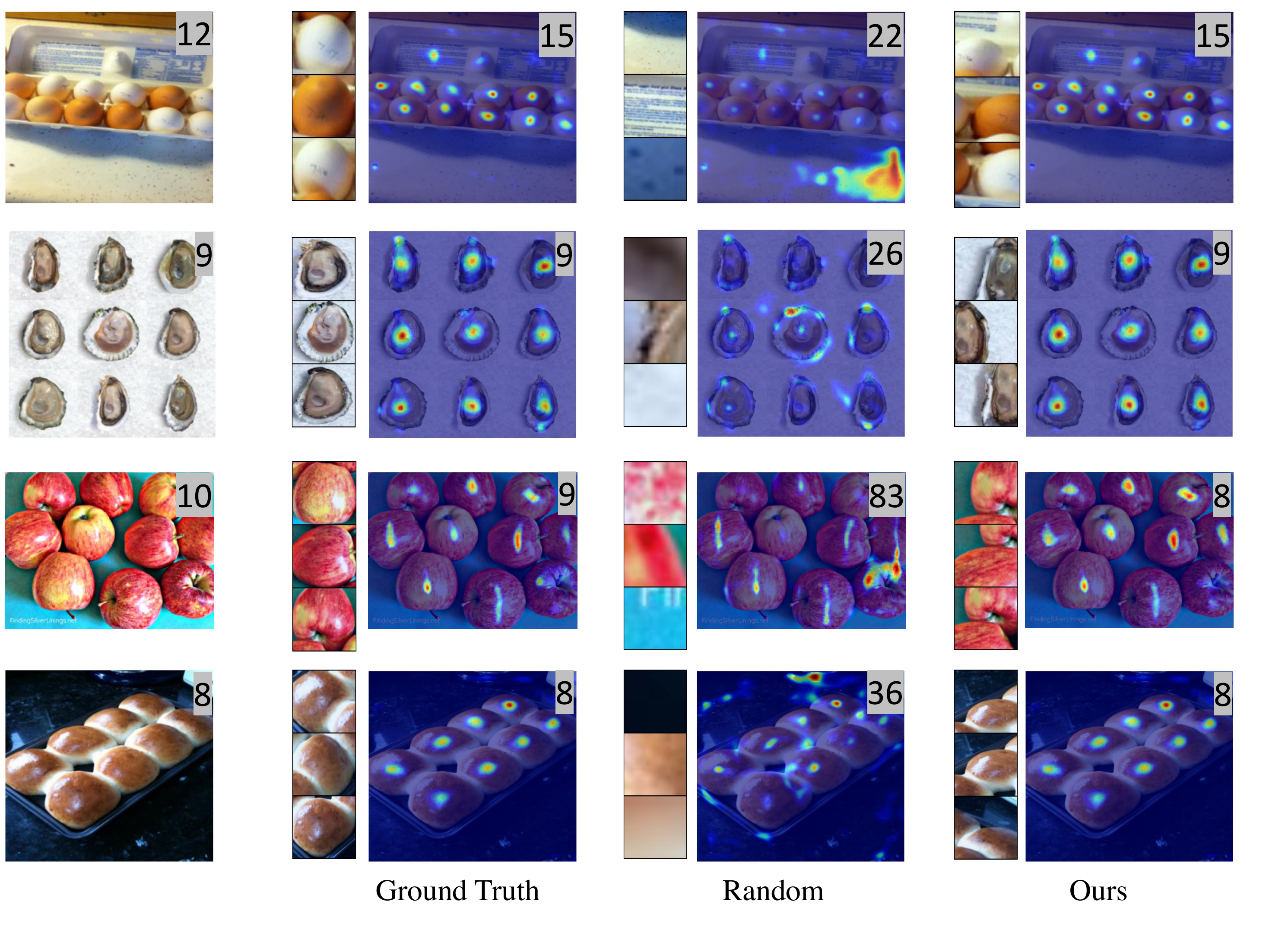}\vspace{-7mm}
 \caption{Qualitative results on the FSC-147 dataset. We show the counting exemplars and the corresponding density maps of ground truth boxes, randomly selected patches, and our selected patches respectively. Predicted counting results are shown at the top-right corner. Our  method  accurately identifies suitable patches for counting and the predicted density maps are close to the ground truth density maps.
} \vspace{-2mm}
\label{fig:img_visualization}
\end{figure*}

\textbf{Qualitative analysis.}  In Figure \ref{fig:img_visualization}, we present a few input images, the image patches selected by our method, and the corresponding density maps. Our method effectively identifies the patches that are suitable for object counting. The density maps produced by our selected patches are meaningful and close to the density maps produced by human-annotated patches. The counting model with random image patches as exemplars, in comparison, fails to output meaningful density maps and infers incorrect object counts.

 \def\subboxsize{0.45\textwidth}

 \def\subboxsize{0.45\textwidth}
 \begin{figure*}[ht!]
 \centering
\includegraphics[width=0.77\linewidth]{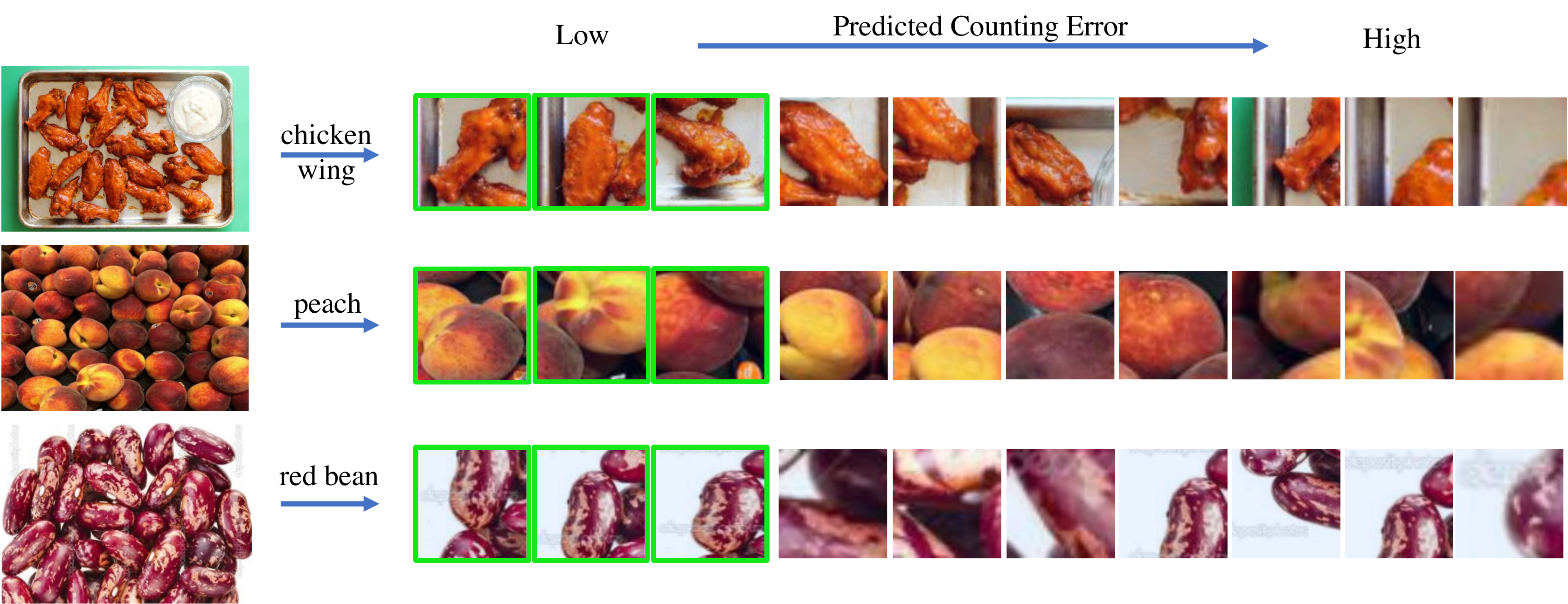} \vspace{-1mm}
 \caption{Qualitative ablation analysis. All the $10$ selected class-relevant patches exhibit some class-specific attributes. They are ranked by the predicted counting errors and the final selected patches  with the smallest errors are framed in green.
}
\label{fig:sel_patches}\vspace{-2mm}
\end{figure*}

\section{Analyses}
\label{sec:ana}
 \subsection{Ablation Studies}
 \label{sec:threshold}
 %See Figure \ref{fig:thres} (Add number of selected features)
Our proposed patch selection method consists of two steps: the selection of class-relevant patches via a generated class prototype and the selection of the optimal patches via an error predictor. We analyze the contribution of each step quantitatively and qualitatively. Quantitative results are in Table \ref{tab:ablation}. We first evaluate the performance of our baseline, i.e. using $3$ randomly sampled patches as exemplars without any selection step. As shown in Table \ref{tab:ablation}, using the class prototype to select class-relevant patches reduces the error rate by $7.19$ and $6.07$ on the validation and test set of MAE, respectively. Applying the error predictor can improve the baseline performance by $7.22$ on the validation MAE and $7.57$ on the test MAE. Finally, applying the two components together further boosts performance, achieving $26.93$ on the validation MAE and $22.09$ on the test MAE. %This validates the effectiveness of each step of our patch selection method.

We  provide further qualitative analysis by visualizing the selected patches. As shown in Figure \ref{fig:sel_patches}, for each input query image, we show $10$ class-relevant patches selected using our generated prototype, ranked by their predicted counting error (from low to high). All the $10$ selected class-relevant patches exhibit some class specific features. However, not all these patches are suitable to be used as counting exemplars, i.e., some patches only contain parts of the object, and some patches contain some background. By further applying our proposed error predictor, we can identify the most suitable patches with the smallest predicted counting errors.

\begin{table}[!h] 
  \centering
\resizebox{0.48\textwidth}{!}{%
  \begin{tabular}{c|c|cccc|cccc}
    \toprule
   \multirow{2}{*}{Prototype}& \multirow{2}{*}{Predictor} & \multicolumn{4}{c|}{Val Set} & \multicolumn{4}{c}{Test Set} \\
    & & MAE & RMSE & NAE & SRE & MAE & RMSE & NAE & SRE \\
    \midrule
    - & - & {35.20} & {106.70} & 0.61 & 6.68 & {31.37} & {134.98} & 0.52 & 5.92 \\
    \Checkmark & - & {28.01} & 88.29 & 0.39 & 4.66 & {25.30} & \textbf{113.82} & 0.40 & 4.88 \\ 
    - & \Checkmark & {27.98} & \textbf{88.62} & 0.43 & 4.59 & {23.80} & {128.36} & 0.40 & 4.43 \\
    \Checkmark & \Checkmark & \textbf{26.93} & {88.63} & \textbf{0.36} & \textbf{4.26} & \textbf{22.09} & {115.17} & \textbf{0.34} & \textbf{3.74} \\
    \bottomrule
  \end{tabular}} \\ %\vspace{-2mm}
  \caption{ Ablation study on each component's contribution to the final results. We show the effectiveness of the two steps of our framework: selecting class-relevant patches via a generated class prototype and selecting optimal patches via an error predictor.
  }\label{tab:ablation} \vspace{0mm}
\end{table}

\subsection{Generalization to Exemplar-based Methods}
\begin{figure}[ht!]
 \centering
\begin{subfigure}{0.48\textwidth}
\hspace{0mm}\includegraphics[width=\linewidth]{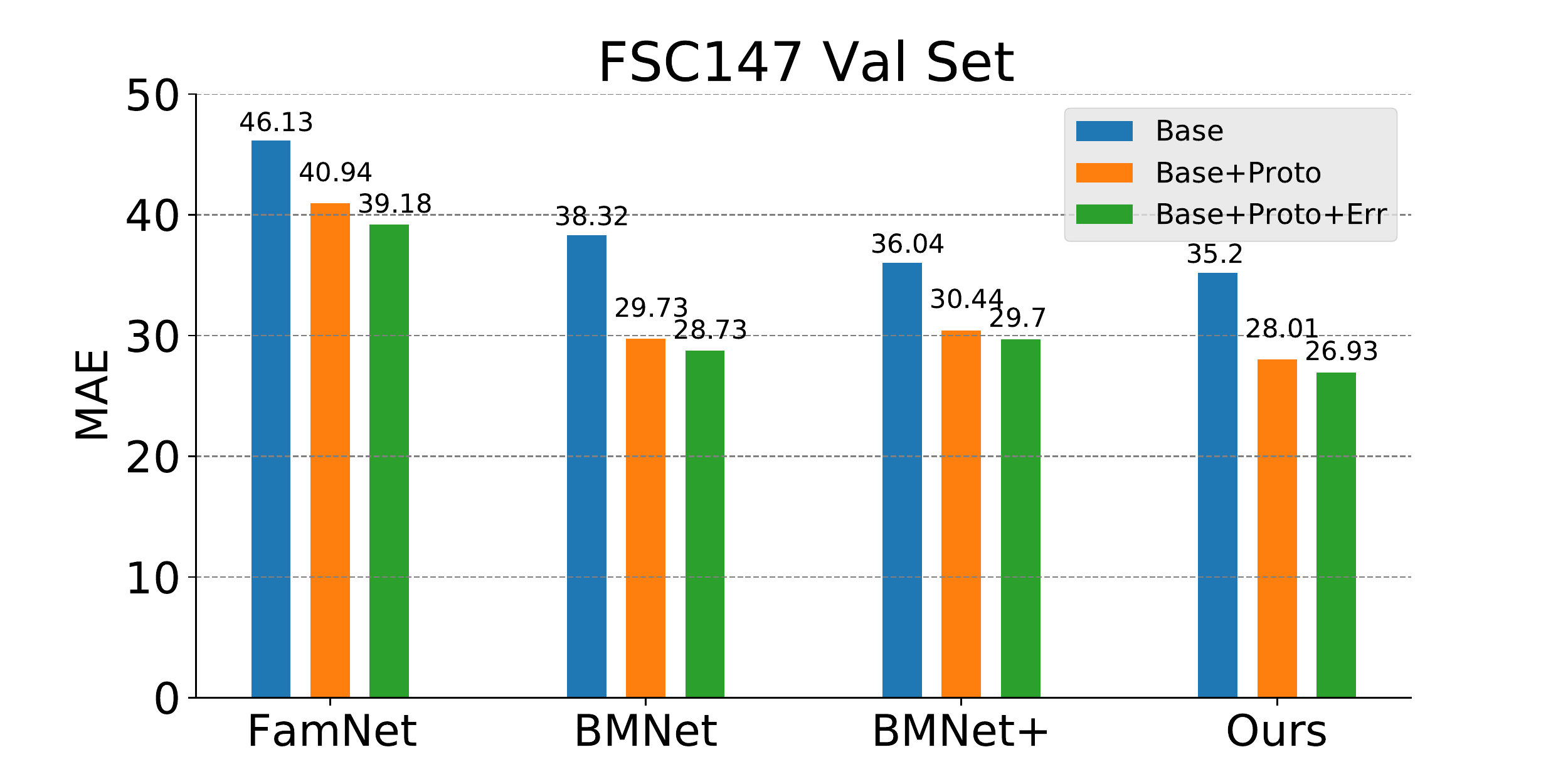} 
\caption{} 
\end{subfigure} 
\begin{subfigure}{0.48\textwidth}
\hspace{0mm}\includegraphics[width=\linewidth]{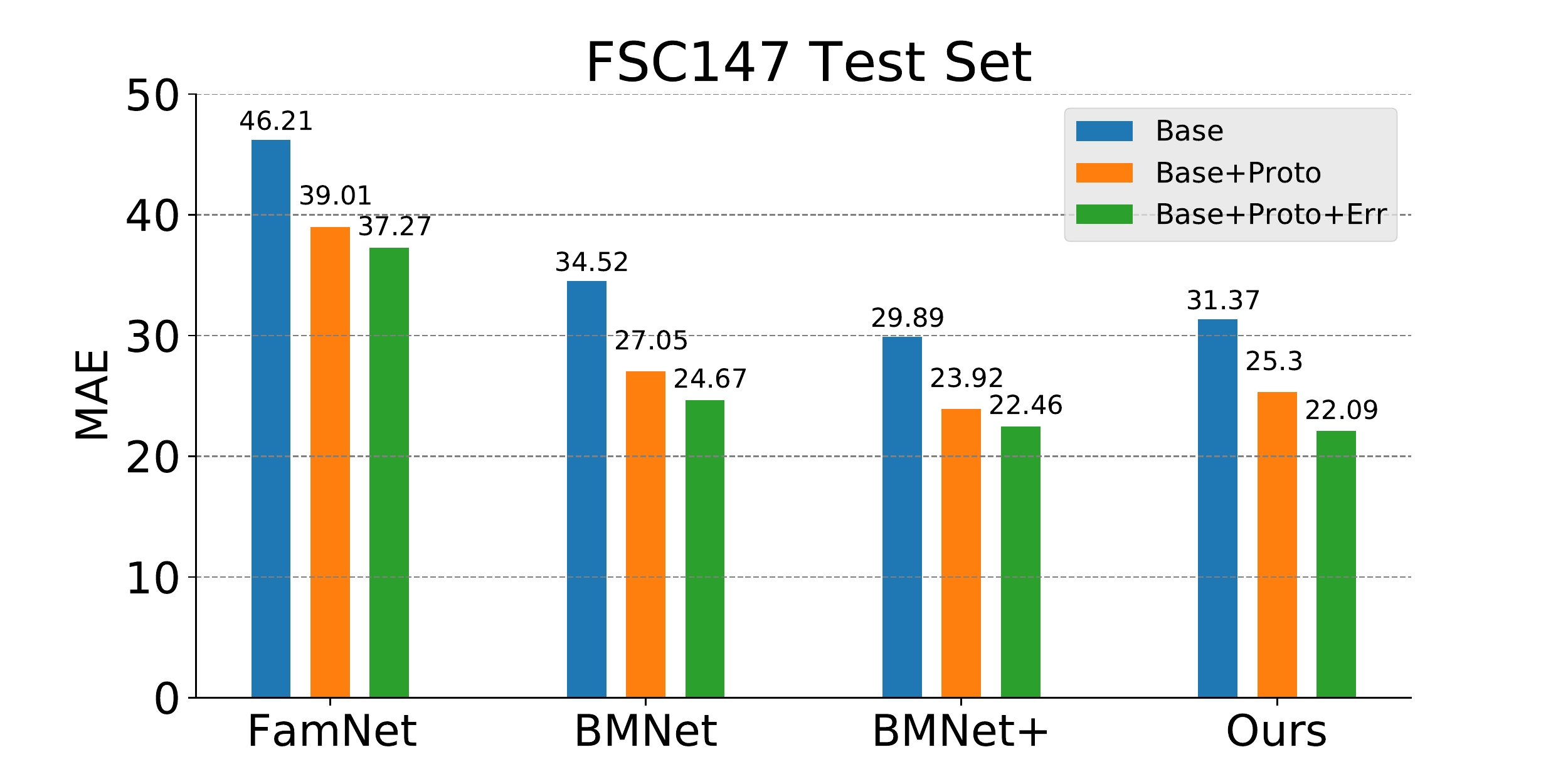}
\caption{}
\end{subfigure} 
 \caption{ {Using our selected patches as exemplars for other exemplar-based class-agnostic counting methods (FamNet, BMNet and BMNet+) on FSC-147 dataset}. Blue bars are the MAEs of using three randomly sampled patches. Orange bars are the MAEs of using the class prototype to select class-relevant patches as exemplars. Green bars are the MAEs of using the class prototype and error predictor to select optimal patches as exemplars. 
} \vspace{-2.5mm}
\label{fig:generalization}
\end{figure}

Our proposed method can be considered as a general patch selection method that is applicable to other visual counters to achieve exemplar-free counting. To verify that, we use our selected patches as the exemplars for three other different exemplar-based methods: FamNet \cite{Ranjan2021LearningTC}, BMNet and BMNet+ \cite{Shi2022SimiCounting}. Figure \ref{fig:generalization} (a) shows the results on the FSC-147 validation set. The baseline uses three randomly sampled patches as the exemplars for the pre-trained exemplar-based counter. By using the generated class prototype to select class-relevant patches, the error rate is reduced by $5.18$, $8.59$ and $5.60$ on FamNet, BMNet and BMNet+, respectively. In addition, as the error predictor is additionally adopted, the error rate is further reduced by $1.76$, $1.00$ and $1.08$ on FamNet, BMNet and BMNet+, respectively. Similarly, Figure \ref{fig:generalization} (b) shows the results on the FSC-147 test set. Our method achieves consistent performance improvements for all three methods. 
 \def\subboxsize{0.36\textwidth}

 \def\subboxsize{0.42\textwidth}
 \begin{figure}[ht!]
 \centering
\begin{subfigure}{0.48\textwidth}
\hspace*{-0.cm}\includegraphics[width=\linewidth]{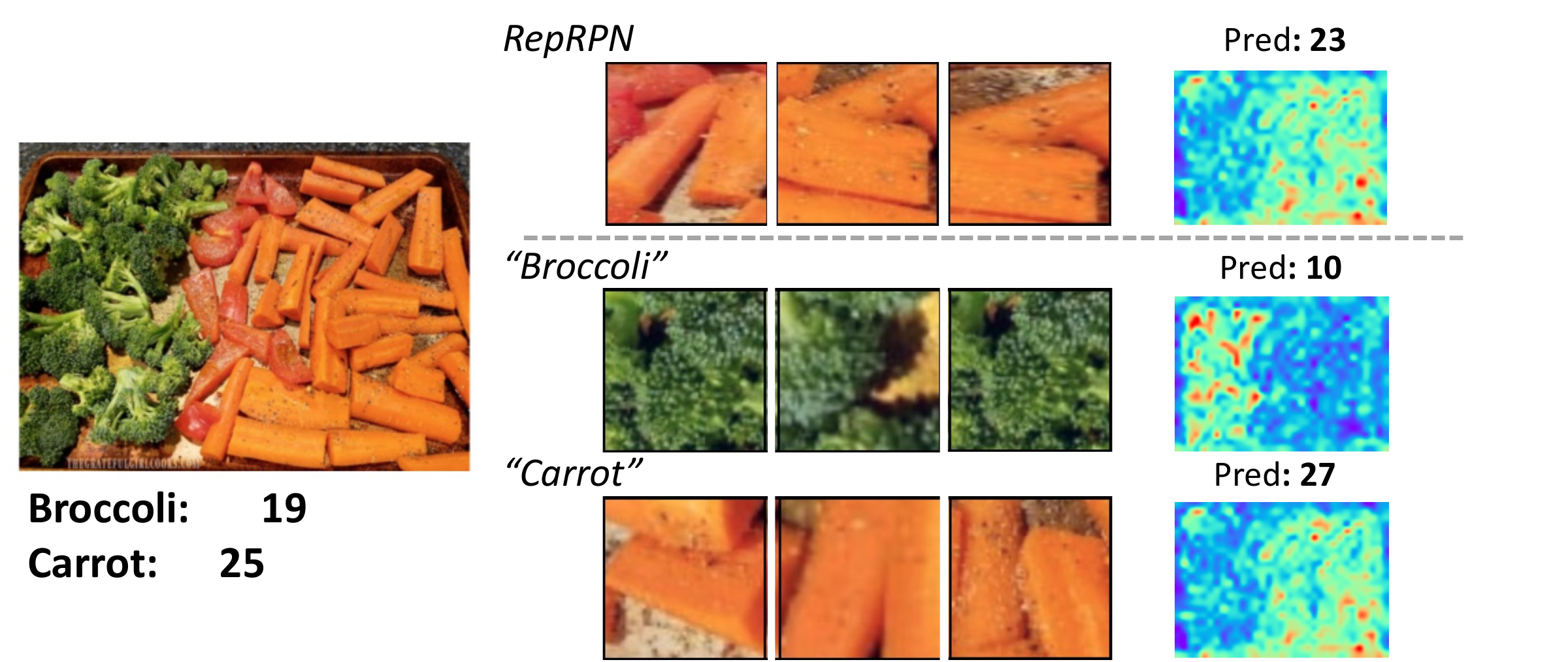}
\caption{}
\label{fig:multi1}
\end{subfigure}
\begin{subfigure}{0.48\textwidth}
\hspace*{-0.cm}\includegraphics[width=\linewidth]{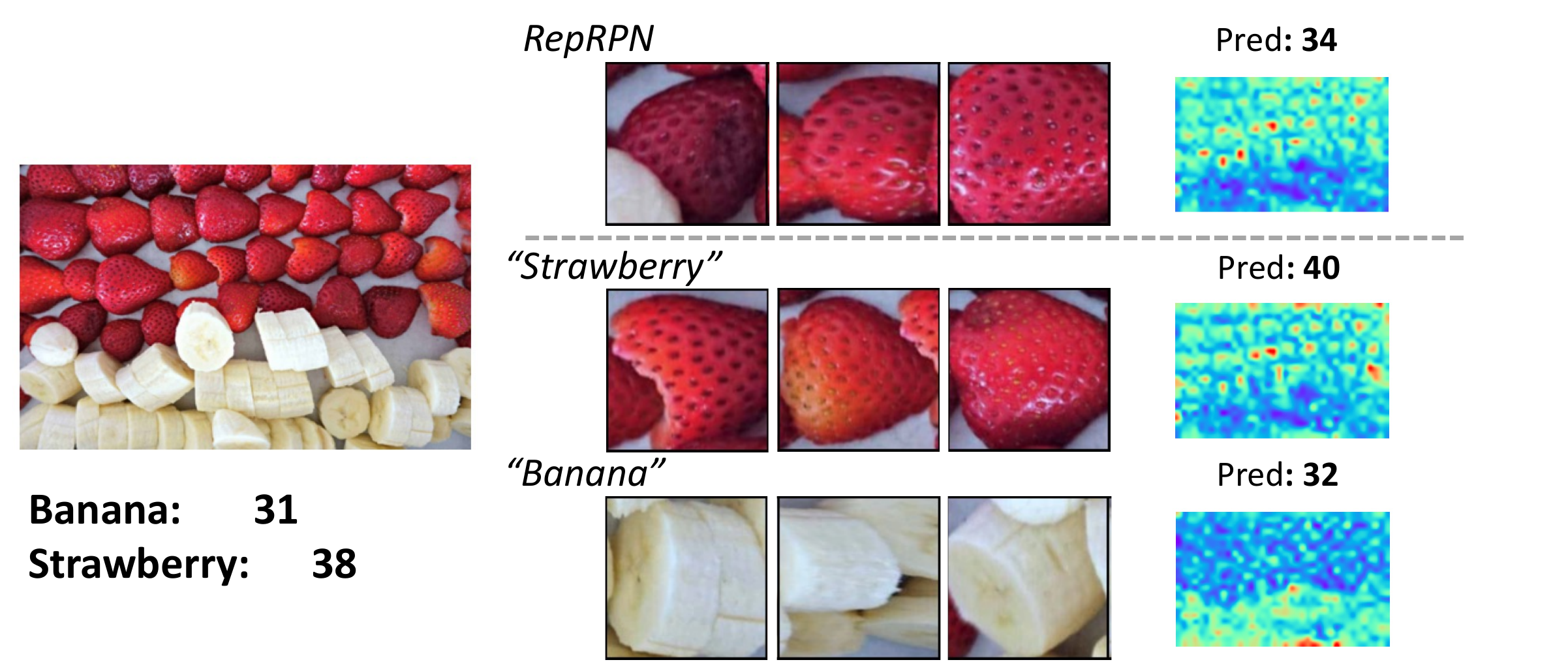}
\caption{}
\label{fig:multi2}
\end{subfigure}
\begin{subfigure}{0.48\textwidth}
\hspace*{-0.cm}\includegraphics[width=\linewidth]{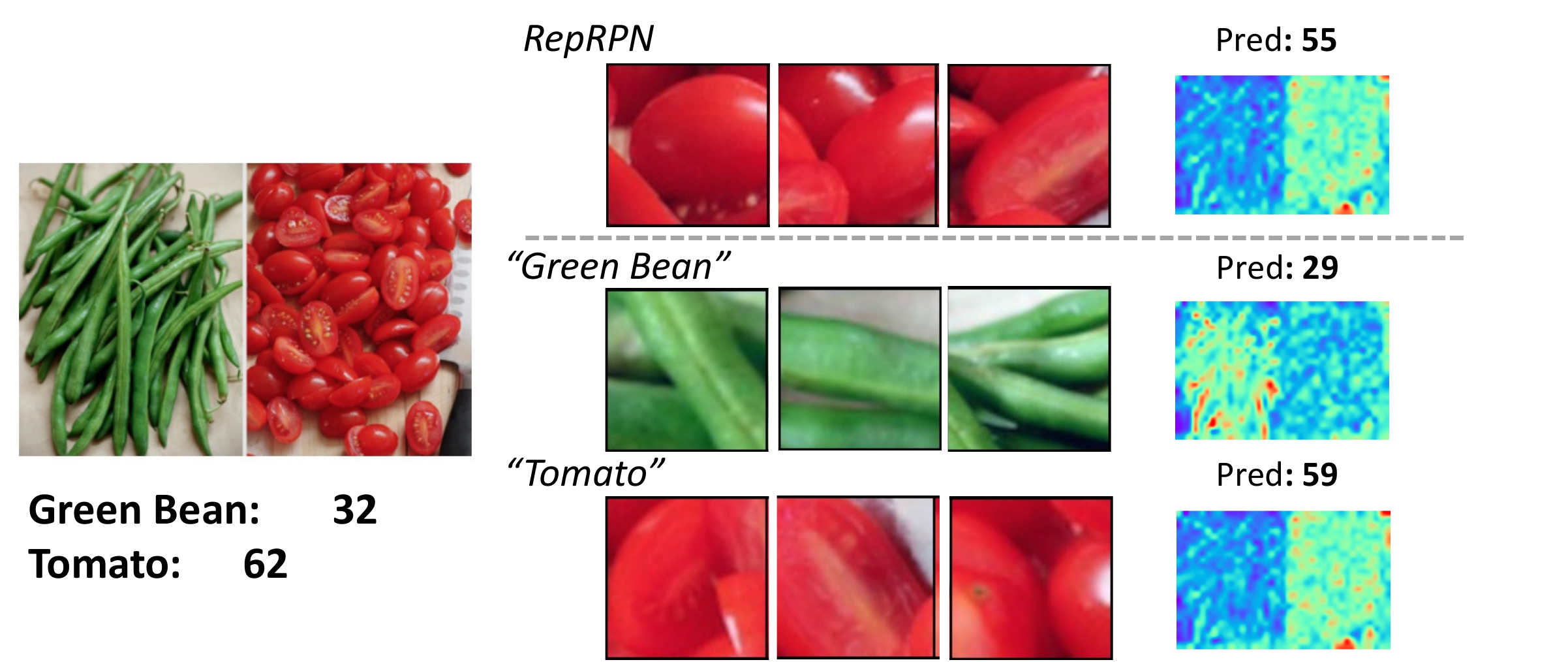} 
\caption{}
\label{fig:multi3}
\end{subfigure} \vspace{-2mm}
\caption{Visualization results of our method in some multi-class examples. Our method  selects patches according to the given class name and the corresponding heatmap  highlights the relevant areas. } 
\label{fig:multi}\vspace{0pt}
\end{figure}

\subsection{Multi-class Object Counting}
Our method can count instances of a specific class given the class name, which is particularly useful when there are multiple classes in the same image. In this section, we show some visualization results in this multi-class scenario. As  seen in Figure \ref{fig:multi}, our method  selects patches according to the given class name and counts instances from that specific class in the input image. Correspondingly, the heatmap highlights the image regions that are most relevant to the specified class. Here the heatmaps are obtained by correlating the exemplar feature vector with the image feature map in a pre-trained ImageNet feature space. Note that we mask out the image region where the activation value in the heatmap is below a threshold when counting.
We also show the patches selected using another exemplar-free counting method, RepRPN \cite{Ranjan2022Exemplar}. The class of RepRPN selected patches can not be explicitly specified. It simply selects patches from the class with the highest number of instances in the image according to the repetition score.
%The selected patches might not belong to the same class (Figure \ref{fig:multi} a) or are not suitable to serve as counting exemplars (Figure \ref{fig:multi} c). 
%For example, in Figure \ref{fig:multi} (a), patches selected using RPN contain objects from different classes. Using these patches as exemplars for counting will lead to inaccurate counting results.

\section{Conclusion} 
In this paper, we proposed a new task, zero-shot object counting, to count instances of a specific class given only the class name without access to any exemplars. To address this, we developed a simple yet effective method that accurately localizes the optimal patches across the query image that can be used as counting exemplars. Specifically, we construct a class prototype in a pre-trained feature space and use the prototype to select patches that contain objects of interest; then we use an error predictor to select those patches with the smallest predicted errors as the final exemplars for counting.
Extensive results demonstrate the effectiveness of our method. We also conduct experiments to show that our selected patches can be used for other exemplar-based counting methods to achieve exemplar-free counting.

\newcommand{\myheading}[1]{\vspace{1ex}\noindent \textbf{#1}}
\myheading{Acknowledgements.} 
This research was partially supported by NSF grants IIS-2123920 and IIS-2212046 and the NASA Biodiversity program (Award 80NSSC21K1027).

\newpage

%%%%%%%%% REFERENCES
%\newpage
\FloatBarrier
{\small
\bibliographystyle{ieee_fullname}
\bibliography{egbib}
}

\end{document}

\title{Zero-Shot Object Counting \\
Supplementary Material}

\author[1]{Jingyi Xu}
\author[2]{Hieu Le}
\author[1]{Vu Nguyen}
\author[3]{Viresh Ranjan\thanks{Work done prior to joining Amazon}}
\author[1]{Dimitris Samaras}
\affil[1]{Stony Brook University} 
\affil[2]{EPFL}
\affil[3]{Amazon}
\maketitle

\section{Overview}
In this document, we provide additional experiments and analyses. In particular:
\begin{itemize}
    \item Section \ref{sec:multi_class} provides additional visualizations of our selected patches in multi-class cases.
    \item Section \ref{sec:candidate_patches} provides the performance of our method when using different sets of candidate patches to select exemplars.
    \item Section \ref{sec:err_score} provides the results of selecting exemplars from the class-relevant patches based on the objectness score.
    \item Section \ref{sec:qual_rpn} provides qualitative comparisons when using RPN proposals as counting exemplars.  
     \item Section \ref{sec:patch_sel} compares our proposed patch selection method with directly using the generated prototype to perform correlation matching to get the  similarity map.
\end{itemize}

\section{Multi-class Zero-shot Counting}
\label{sec:multi_class}
\begin{figure*}[!ht]
\begin{center}
\hspace*{0.48cm}\includegraphics[width=0.9\columnwidth]{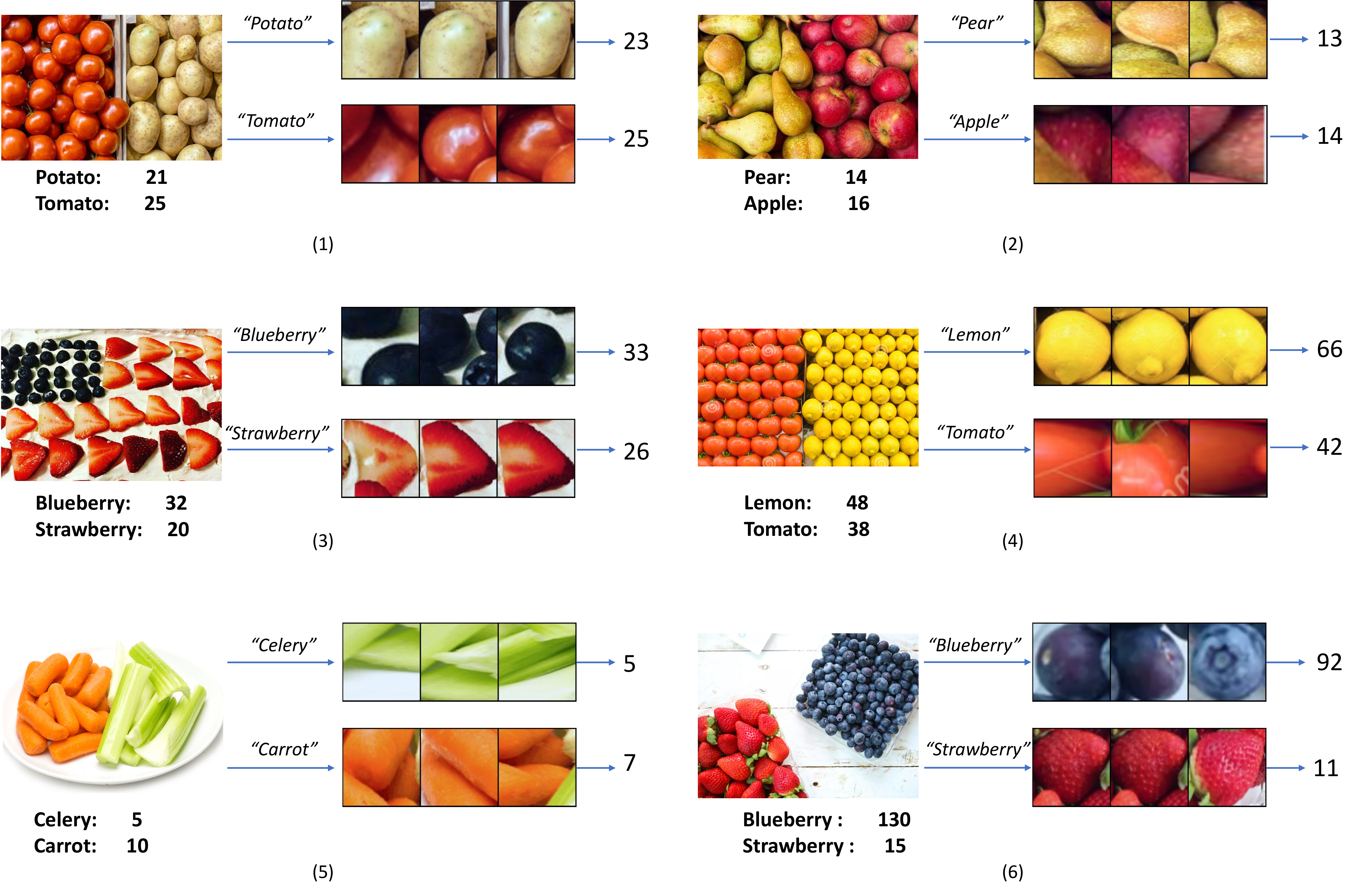}

\caption{ Visualizations of the selected patches: There are two classes with multiple object instances in a single image. To specify what to count, we provide the class name at test time. Our proposed method selects counting exemplars according to the given class name and count instances from the specific class.
} 
\label{fig:multi}
\end{center}

\end{figure*}

Figure \ref{fig:multi} provides additional visualizations of the selected patches in multi-class cases. %Note that images in the FSC-147 dataset contain a single dominant class per image. For images with multiple classes, the previous exemplar-free method, RepRPN \cite{Ranjan2021LearningTC}, is not applicable. 
As can be seen from the figure, our proposed method can select counting exemplars according to the given class name and count instances from that specific class in the input image.

 \section{Different Methods to Acquire Candidate Patches}
\label{sec:candidate_patches} 
In our main experiments, the candidate patches for selection are randomly sampled from the input image. In this section, we conduct an ablation study on how to obtain the candidate patches. Instead of using randomly sampled patches, we take the proposals generated by RPN as the candidate patches and apply our patch selection method. We further combine the random patches and RPN generated proposals together as the candidate patches and evaluate the performance. Results are summarized in Table \ref{tab:sampling}. As can be seen from the table, our proposed patch selection method can bring consistent performance improvements for all the three set of candidate patches. 

\begin{table*}[!ht] 
  \centering
\resizebox{0.75\textwidth}{!}{%
  \begin{tabular}{c|c|cccc|cccc}
    \toprule
   \multirow{2}{*}{\parbox{1.5cm}{Candidate \\ Patches}} & \multirow{2}{*}{Patch Selection} & \multicolumn{4}{c|}{Val Set} & \multicolumn{4}{c}{Test Set} \\
   & & MAE & RMSE & NAE & SRE & MAE & RMSE & NAE & SRE \\
    \midrule
    \multirow{2}{*}{RPN} &  \xmark & 29.64 & 94.93 & 0.43 & 4.96 & 25.04 & 126.02 & 0.36 & 4.18  \\ 
       & \checkmark & 26.76 & 87.41 & \textbf{0.34} & 4.36 & 23.52 & 126.68 & 0.33 & 3.94
 \\
       \midrule
     \multirow{2}{*}{Random} & \xmark & 35.20 & 106.70 & 0.61 & 6.68 & 31.37 & 134.98 & 0.52 & 5.92 \\
      & \checkmark & 26.93 & 88.63 & 0.36 & \textbf{4.26} & {22.09} & \textbf{115.17} & {0.34} & {3.74} \\
     \midrule
     \multirow{2}{*}{Random+RPN} & \xmark & {29.97} & {91.61} & {0.44} & {5.17} & {24.91} & {126.35} & {0.36} & {4.45} \\
      & \checkmark & \textbf{26.58} & \textbf{86.69} & 0.35 & 4.28 & \textbf{22.03} & 116.42 & \textbf{0.33} & \textbf{3.65} \\
    \bottomrule
  \end{tabular}} \\ \vspace{6pt}
  \caption{ Performance on FSC-147 dataset when using different sets of candidate patches to do patch selection. Our proposed method brings consistent improvement in the performance.
  }\label{tab:sampling}%
\end{table*}
 
 \section{Comparing Predicted Errors with Objectness Scores}
 \label{sec:err_score}

%We have shown that replacing human-annotated exemplars with RPN proposals will lead to a significant performance drop in our main experiments. 
%Specifically, we first use the generated prototype to select $10$ class-relevant patches. Then we compare the performance of using our proposed error predictor to select among these $10$ patches and using the objectness score from RPN. 
 In our proposed method, after obtaining the class-relevant patches, we select among them the final counting exemplars via an error predictor.
 To further validate the effectiveness of the error predictor, we compare the performance of our method with a baseline approach which simply uses the objectness score from the RPN to select counting exemplars from the class-relevant patches. 
 Specifically, after obtaining the class-relevant patches, we rank them according to the objectness score and select the patches with the top-$3$ highest objectness scores as counting exemplars. As shown in Table \ref{tab:err_pred}, our method outperforms the baseline using the objectness score in most cases, which shows the effectiveness of our proposed error predictor.

\begin{table*}[!h] 
  \centering
\resizebox{0.6\textwidth}{!}{%
  \begin{tabular}{c|cccc|cccc}
    \toprule
   \multirow{2}{*}{} & \multicolumn{4}{c|}{Val Set} & \multicolumn{4}{c}{Test Set} \\
    & MAE & RMSE & NAE & SRE & MAE & RMSE & NAE & SRE \\
    \midrule
   Obj Score & {28.47} & {94.87} & \textbf{0.34} & 4.65 & {24.11} & {117.76} & 0.35 & 4.00 \\ 
     Pred Error & \textbf{26.93} & \textbf{88.63} & 0.36 & \textbf{4.26} & \textbf{{22.09}} & \textbf{{115.17}} & \textbf{0.34} & \textbf{3.74} \\
    \bottomrule
  \end{tabular}} \\ \vspace{6pt}
  \caption{ {Comparison between using the predicted counting error and the objectness score from RPN to select among class-relevant patches.} 
  }\label{tab:err_pred}%

\end{table*}

\section{Qualitative Comparison with RPN} 
\label{sec:qual_rpn}
\begin{figure*}[!h]
\begin{center}
\hspace*{0.48cm}\includegraphics[width=0.68\columnwidth]{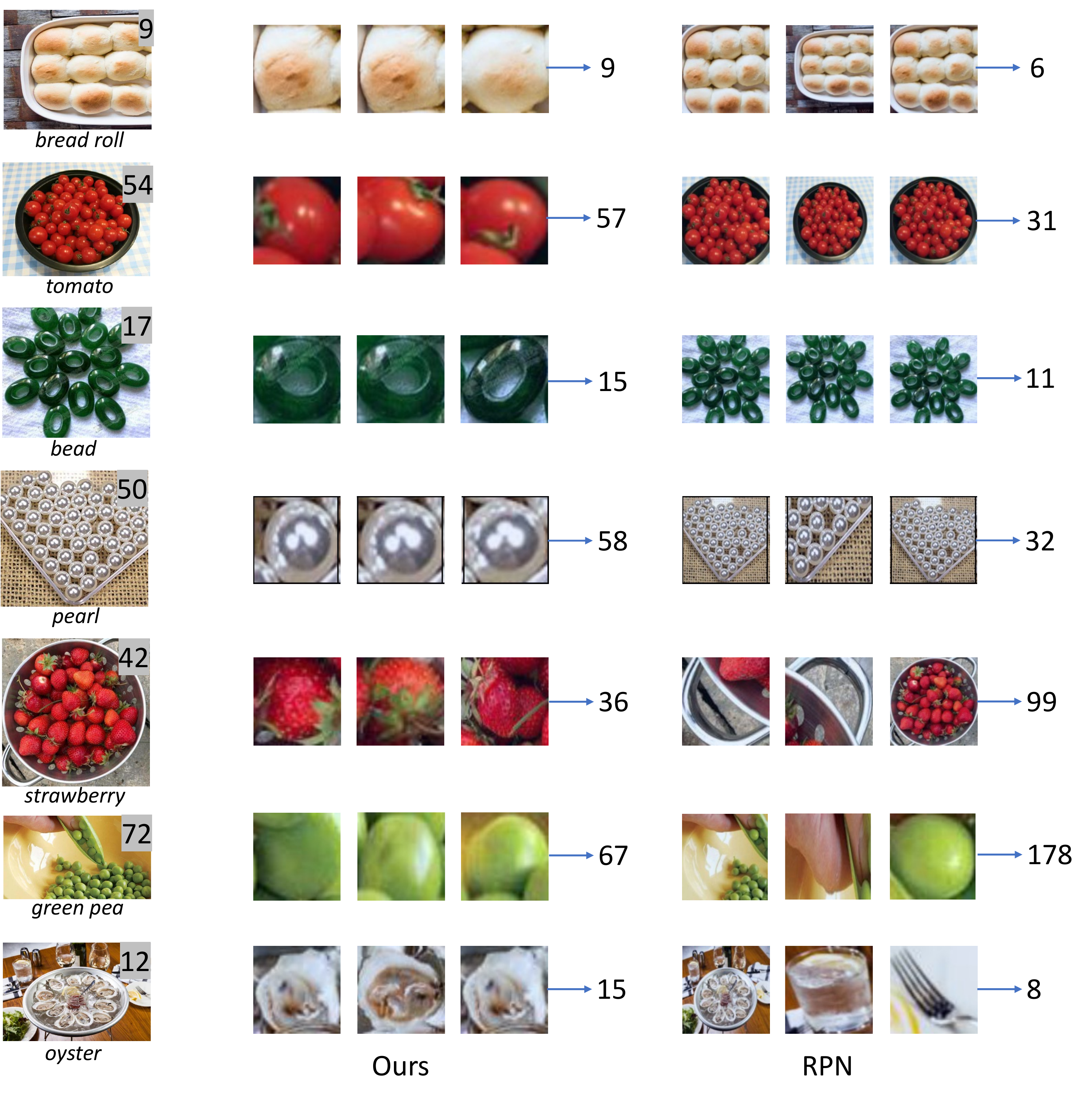}
\vspace{-2mm}
\caption{ Qualitative comparison with RPN. Our proposed method can select patches suitable for counting while RPN-selected patches contain non-relevant objects or multiple instances.
} 
\label{fig:rpn}\vspace{-9mm}
\end{center}

\end{figure*}

In Figure \ref{fig:rpn}, we visualize some images from the FSC-147 dataset and the corresponding patches selected by our proposed method and RPN, respectively. The RPN-selected patches are the top-3 proposals with the highest objectness scores. As can be seen from the figure, our proposed method can accurately localize image patches according to the given class name. These selected patches can then be used as counting exemplars and yield reasonable counting results. In comparison, the patches selected by RPN might contain objects which are not relevant to the provided class name or contain multiple instances. These patches are not suitable to be used as counting exemplars and will lead to inaccurate counting results. This suggests that choosing counting exemplars based on objectness score is not reliable.

\section{Comparing with Correlation Matching via Prototype}
\label{sec:patch_sel} 
 Our strategy for zero-shot counting is to select patches across the query image and use them as exemplars for counting. The patches are selected via a generated class prototype and an error predictor. An alternative way is to use the generated prototype to do correlation matching directly instead of selecting patches from the input image. In this section, we compare the performance of these two strategies and show the advantage of our proposed one. Specifically, we use the generated prototype to do correlation matching with the features of input images to get the similarity map, which will then be given as input to the counter to get the density map and final count. The VAE used to generate the prototype in our main experiments is trained on the MS-COCO detection set. We train another VAE using the FSC-147 training set and report the performance with both VAEs in Table \ref{tab:patch_selection}. As can be seen, our proposed patch selection method achieves better results. Directly using the generated prototype to perform correlation matching provides a simple solution for zero-shot counting. However, it is not optimal since the same prototype is applied to all the objects from different images. These objects typically exhibit large variability. Our patch selection method, in comparison, selects different exemplars dynamically according to the input image. %The selected exemplars are more likely to share fine-grained details with the objects in the image compared with a universal prototype.
\begin{table*}[!h] 
  \centering
\resizebox{0.7\textwidth}{!}{%
  \begin{tabular}{c|c|cccc|cccc}
    \toprule
   \multirow{2}{*}{\parbox{1.5cm}{Patch \\ selection}}& \multirow{2}{*}{\parbox{2cm}{Training data\\for VAE}} & \multicolumn{4}{c|}{Val Set} & \multicolumn{4}{c}{Test Set} \\
    & & MAE & RMSE & NAE & SRE & MAE & RMSE & NAE & SRE \\
    \midrule
    \xmark & MS-COCO & {48.56} & {127.93} & 0.65 & 6.37 & {41.33} & {147.43} & 0.52 & 5.53 \\ 
    \xmark & FSC-147 & {30.51} & {101.39} & 0.41 & 4.66 & {28.03} & {132.34} & 0.37 & 4.42 \\
    \Checkmark & MS-COCO & \textbf{27.00} & \textbf{87.90} & \textbf{0.35} & \textbf{4.29} & \textbf{22.09} & \textbf{115.17} & \textbf{0.34} & \textbf{3.74} \\
    \bottomrule
  \end{tabular}} \\ \vspace{4pt}
  \caption{ {Comparison between our proposed method and the baseline approach of directly using the generated class prototype to do correlation matching.} 
  }\label{tab:patch_selection}%
  \vspace{-10pt}
\end{table*}
  
\section{Qualitative Analysis}
\label{sec:qual_ana} 
\begin{figure*}[!h]
\begin{center}
\hspace*{0.34cm}\includegraphics[width=0.47\columnwidth]{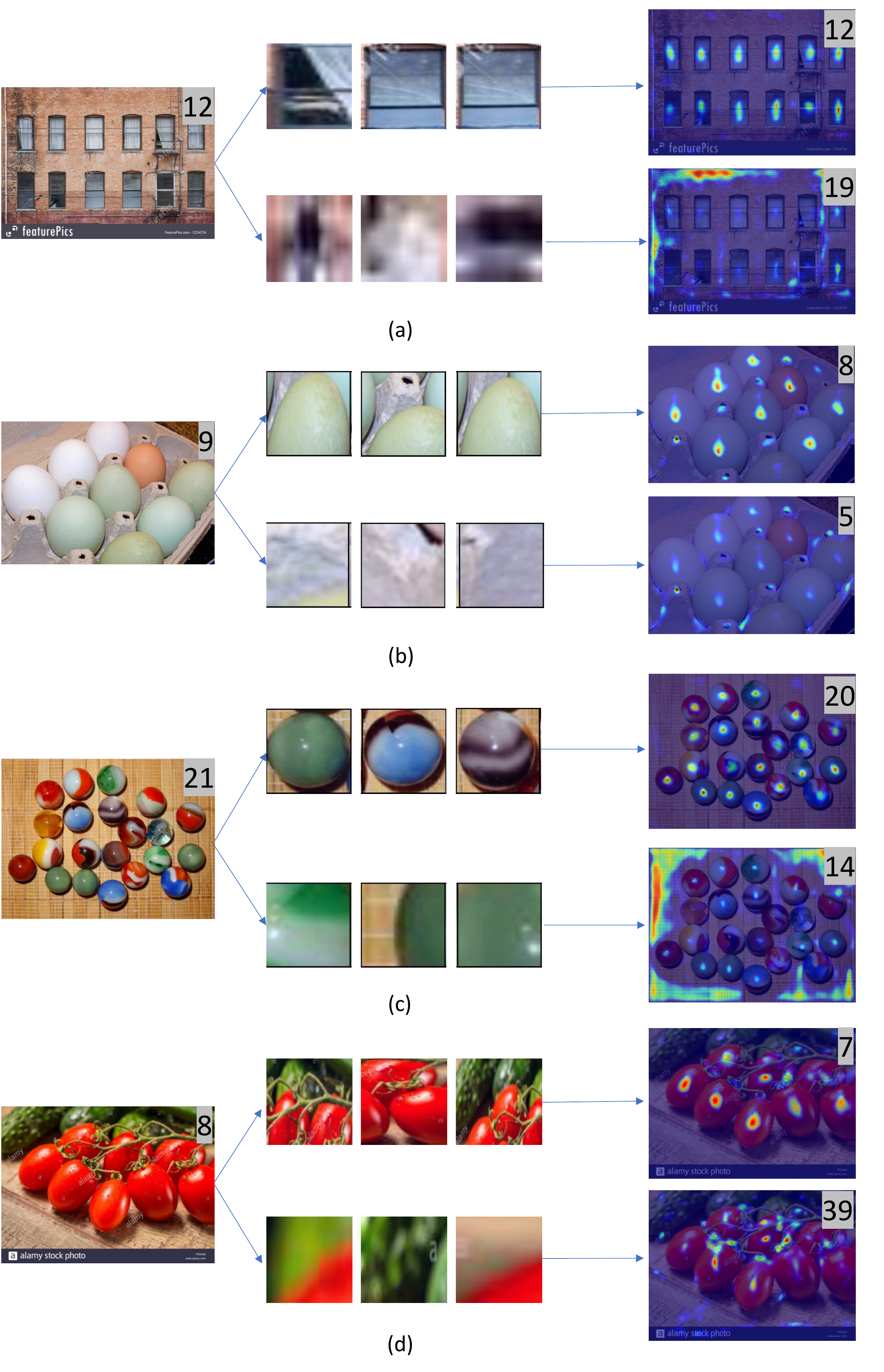}
\caption{ Visualizations of the patches with top-$3$ lowest and highest predicted counting errors.
} 
\label{fig:map}
\end{center}

\end{figure*}

In Figure \ref{fig:map}, we show a few input images and the corresponding patches with the  top-$3$ lowest and highest predicted counting errors. As can be seen from the figure, the patches with the smallest predicted errors are suitable to serve as counting exemplars and output meaningful density maps and accurate counting results. In comparison, the density maps produced by patches with the highest predicted errors fail to highlight the relevant image regions and lead to inaccurate counting results. This suggests that the predicted counting error can effectively indicate the goodness of the counting exemplars.

%%%%%%%%% REFERENCES
\newpage
%\FloatBarrier
%{\small
%\bibliographystyle{ieee_fullname}
%\bibliography{egbib}
%}